\documentclass[10pt,twocolumn,letterpaper]{article}

\usepackage[pagenumbers]{cvpr} 

\usepackage{graphicx}
\usepackage{amsmath}
\usepackage{amssymb}
\usepackage{booktabs}
\usepackage{url}

\usepackage{utfsym}
\usepackage{pifont}
\usepackage{soul} 
\usepackage{dsfont}

\usepackage[pagebackref,breaklinks,colorlinks]{hyperref}

\usepackage{multirow}
\usepackage{pifont}
\usepackage{colortbl}
\usepackage[capitalize,noabbrev]{cleveref}
\usepackage{mathrsfs}
\crefname{section}{Sec.}{Secs.}
\Crefname{section}{Section}{Sections}
\Crefname{table}{Table}{Tables}
\crefname{table}{Tab.}{Tabs.}

\makeatletter
\newcommand{\thickhline}{%
	\noalign {\ifnum 0=`}\fi \hrule height 1pt
	\futurelet \reserved@a \@xhline
}
\makeatother

\def\cvprPaperID{*****} 
\def\confName{CVPR}
\def\confYear{2024}

\begin{document}

\title{The Instance-centric Transformer for the RVOS Track of LSVOS Challenge: \\ 3\textsuperscript{rd} Place Solution}  

\author{Bin Cao$^{1,2,3}$\thanks{Equal contribution.}{\qquad} Yisi Zhang$^{4}${\qquad} Hanyi Wang$^{2}${\qquad}Xingjian He$^{1}${\qquad}Jing Liu$^{1,2}$\thanks{ Corresponding author.}
\vspace{3mm}\\
$^{1}$Institute of Automation, Chinese Academy of Sciences (CASIA)\qquad \\
$^{2}$School of Artificial Intelligence, University of Chinese Academy of Sciences (UCAS) \\
$^{3}$Beijing Academy of Artificial Intelligence (BAAI) \\
$^{4}$University of Science and Technology Beijing (USTB) \\
Team: CASIA\_IVA
}

\maketitle
\begin{abstract}
Referring Video Object Segmentation is an emerging multi-modal task that aims to segment objects in the video given a natural language expression. In this work, we build two instance-centric models and fuse predicted results from frame-level and instance-level. First, we introduce instance mask into the DETR-based model for query initialization to achieve temporal enhancement and employ SAM for spatial refinement. Secondly, we build an instance retrieval model conducting binary instance mask classification whether the instance is referred. Finally, we fuse predicted results and our method achieved a score of 52.67 $\mathcal{J}\&\mathcal{F}$ in the validation phase and 60.36 $\mathcal{J}\&\mathcal{F}$ in the test phase, securing the final ranking of 3rd place in the 6-th LSVOS Challenge RVOS Track.
\end{abstract}

\section{Introduction}
Referring Video Object Segmentation (ROVS) aims to segment and track the target object referred by the given language description.
This emerging task has attracted significant attention due to its potential applications in video editing and human-agent interaction.

There are two tracks of the LSVOS Challenge this year: Referring Video Object Segmentation (RVOS) Track and Video Object Segmentation (VOS) Track. Upgraded from the Refer-Youtube-VOS dataset, this year's RVOS track features the newly proposed MeViS\cite{ding2023mevis} dataset. Compared to the conventional RVOS datasets like Ref-Youtube-VOS \cite{seo2020urvos} and Ref-DAVIS17 \cite{khoreva2019video}, MeViS presents more complex expressions that include motion information rather than merely simple spatial location descriptions. Given videos and motion-oriented language expressions obtained from MeViS \cite{ding2023mevis}, an embodied agent is needed to segment targets. Consequently, MeViS necessitates that the agent comprehends both temporal and spatial information within video clips to effectively correlate with motion expressions. Furthermore, MeViS extends the RVOS to include language expressions that match multiple targets, making MeViS more challenging and reflective of real-world scenarios. 

With the development of deep learning, there are studies dealing with the RVOS task. For example, some studies \cite{seo2020urvos,bellver2023closer} try to deal with the task from a per-frame perspective, they transfer the referring image segmentation methods \cite{huang2020referring, luo2020multi, ding2021vision} into video domain, whether through a per-frame mask propagation manner or based on history memory attention to predict the mask of the current frame. Most recent methods, e.g. \cite{wu2022language,miao2023spectrum,luo2024soc,yan2024referred,wu2023onlinerefer,botach2022end}, focus on a unified framework that employs language as queries to segment and track the referred object simultaneously. By effectively building correlations between expressions and multi-frame visual features, they achieve promising results across multiple RVOS benchmarks. Some recent methods, e.g.\cite{yan2023universal,wu2023general,li2024univs}, unify various kinds of object-level tasks, such as MOT, VIS, RVOS and RES, into a single framework to present an object-centric foundation model. However, these methods still encounter the issue of inconsistent predicted results across multiple frames. While some recent studies on Video Instance Segmentation (VIS) task\cite{yang2019video}, which emphasizes segmenting different instances in the given video, have shown promising results in dealing with prediction inconsistent problem. Furthermore, the emergence of SAM \cite{kirillov2023segment} also provides strong segmentation tools for refinement. 

\begin{figure*}[t]
  \centering
   \includegraphics[width=1.0\linewidth]{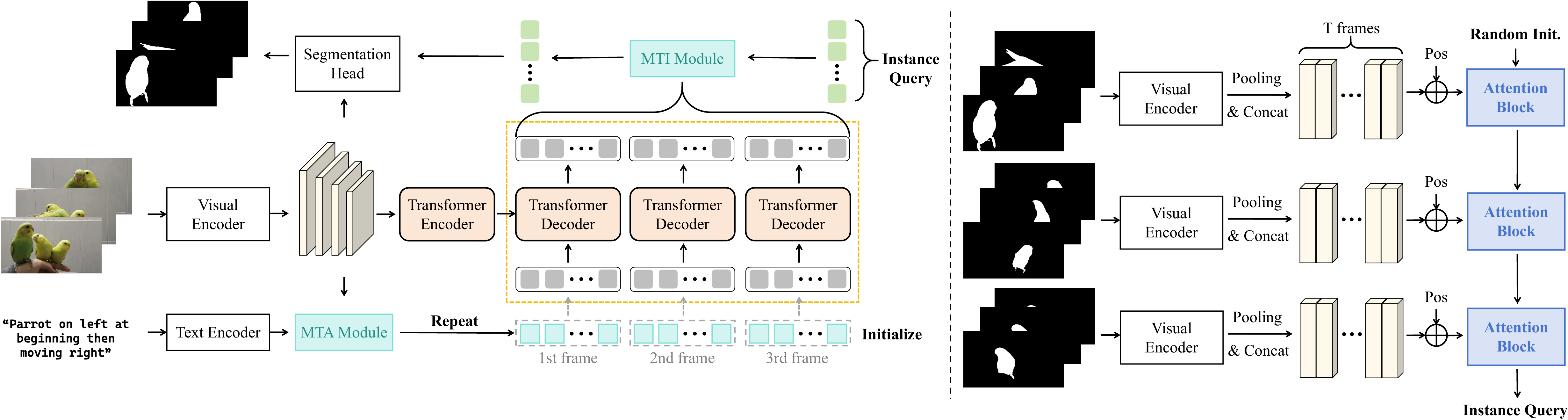}
   \caption{\textbf{The architecture of MUTR-based model.} We employ MUTR as our basic model (\textbf{Left}). We introduce instance masks and employ an attention block and a sequential mechanism to aggregate instance information into a query (\textbf{Right}).}
   \label{fig:model_architecture}
\end{figure*}

Thanks to the superior performance of DVIS \cite{zhang2023dvis}, MUTR \cite{yan2024referred} and HQ-SAM \cite{ke2024segment}, our method achieves a score of 52.67 $\mathcal{J}\&\mathcal{F}$ in the validation phase and 60.36 $\mathcal{J}\&\mathcal{F}$ in the test phase, securing the final ranking of 3rd place in the 6-th LSVOS Challenge RVOS Track.
\label{sec:intro}

\section{Method}
\subsection{Overview}
Our solution contains three components: MUTR-based model; instance retrieval model and fusion strategy. The architecture of MUTR-based solution is shown in \cref{fig:model_architecture}. To improve the consistency of results, we introduce proposal instance masks into MUTR for query initialization. 

\subsection{MUTR-based Model}
MUTR was proposed in \cite{yan2024referred} and has shown superior performance on Ref-Youtube-VOS. MUTR adopts a DETR-like style model. Compared with other methods, MUTR introduces two core modules, i.e. MTI, MTA.  

While MUTR achieves superior performance on RVOS, prediction results from MUTR still suffer from inconsistency and incompleteness. Meantime, some recent studies on VIS show promising results to solve this issue. Therefore, we attempt to introduce instance mask information into a DETR-based model to improve the consistency and completeness of prediction results.

Specifically, we attempt to introduce instance masks to initialize the video-wise query $\mathcal{Q}$ in MTI decoder. Thanks to the superior performance of DVIS on VIS, we employ DVIS for mask generation, which extracts all instance masks in a video clip as follows:
\begin{equation}
    m_{i} = \text{DVIS}(\text{I}), \ m_{i} \in \mathbb{R}^{\ T \times H \times W}
\end{equation}
where $\text{I} \in \mathbb{R}^{\ T \times H \times W \times \text{3}}$ is the input video clip, $m = \{m_{i}\}^{K}_{i=1}$ denotes the set of instance masks, $K$ is the number of instances in a video clip and $T$ is the number of frames.

The motion property is a significant aspect that can distinguish different objects. Therefore, we inject motion cues into instance features. Given a multi-frame instance binary mask $m_{i}$, we calculate the bounding box of this object for each frame and obtain the positional information as follows:
\begin{equation}
   p_{i,t} = (x^{i,t}_{min}, y^{i,t}_{min}, x^{i,t}_{max}, y^{i,t}_{max}, x^{i,t}_{c}, y^{i,t}_{c}, w_{i,t}, h_{i,t})
\end{equation}
where $(x^{i,t}_{min}, y^{i,t}_{min}), (x^{i,t}_{max}, y^{i,t}_{max}), (x^{i,t}_{c}, y^{i,t}_{c}), w_{i,t}$ $h_{i,t}$ are normalized top-left coordinates, bottom-right coordinates, center coordinates, width and height of bounding box respectively, $t$ is the index of video frames.

Next, we utilize a visual encoder to extract multi-scale visual features of instance masks and inject the instance trajectory into visual features as follows:
\begin{equation}
    \mathcal{F}_{i,j} = \text{Visual\_Backbone}(m_{i}) + W(p_{i}), \ \mathcal{F}_{i,j} \in \mathbb{R}^{\ T \times h_{j} \times w_{j} \times c_{j} }
\end{equation}
where $c_{j}$ is the channel of $j$ level visual feature and $W$ is a linear layer. After feature extraction, we utilize a projection layer on multi-scale visual features to align dimension with video features and perform average pooling along spatial dimension to obtain instance features as follows: 
\begin{equation}
    \mathcal{F}^{'}_{i,j} = \text{Pooling}(\text{Proj}(\mathcal{F}_{i,j})), \  \mathcal{F}_{i,j} \in \mathbb{R}^{\ T \times C }
\end{equation}

For simplicity, we only explain our solution utilizing the single-level visual feature. To aggregate all instance information into an instance query, we design an attention block and adapt sequential mechanisms as follows: 
\begin{equation}
    \mathcal{Q}_{i} = \text{Block}(\mathcal{Q}_{i-1},\mathcal{F}^{'}_{i}), \ 1 \leq i \leq K 
\end{equation}
where ${Q}_{i} \in \mathbb{R}^{\ N \times C}$ is the instance query and $N$ is the number of queries. ${Q}_{0}$ is randomly initialized. The designed attention block consists of a cross-attention layer, a set of self-attention layers, and FFN layers. After that, we utilize this query with instance information to replace the randomly initialized video-wise query fed to MTI decoder.

\begin{table*}[ht]
\centering
\caption{\textbf{Ablation Experiment Results on MeViS Validation Set about MUTR-based Model.} We ablate multiple design choices, including sampling method, HQ-SAM, and introducing instance masks for query initialization. In all experiments, we adopt the pre-trained weights of MUTR as initialization and fine-tune model on MeViS.}
\label{tab:ablation}
\vspace{5pt}
\resizebox{1.0\linewidth}{!}{
\begin{tabular}{c|c|c|c|c|c|ccc}
\toprule[1.5pt]
Method  &Backbone & Sampling Method  & Instance Masks  &Position Embedding & HQ-SAM    & $\mathcal{J}\&\mathcal{F}$   & $\mathcal{J}$   & $\mathcal{F}$      \\ 
\midrule
\multirow{7}{*}{MUTR}   & \multirow{7}{*}{Swin-L}                           
& \multirow{2}{*}{Local Sampling}  &\ding{55}  &\ding{55} &\ding{55} &48.40   &44.87  &51.94 
\\
& &  &\ding{55} &\ding{55} & {\checkmark} & 48.66  & 45.86 & 51.46    \\ 
\cmidrule(lr){3-9}
& & \multirow{2}{*}{Global Sampling}  &\ding{55}  &\ding{55} &\ding{55} &49.11   &45.89  &52.33 
\\
& &  &\ding{55} &\ding{55} & {\checkmark} & 49.50  & 46.91 & 52.09    \\ 
\cmidrule(lr){3-9}
& & \multirow{3}{*}{Global Sampling}  & {\checkmark}  &\ding{55} &\ding{55} &49.62   &46.38  &52.85 
\\
& & & {\checkmark} &\ding{55} & {\checkmark} &49.92  & 47.30 & 52.54    \\ 

& & & {\checkmark} & {\checkmark} & {\checkmark} & \cellcolor{gray!25}\textbf{50.27}  & \cellcolor{gray!25}\textbf{47.67} & \cellcolor{gray!25}\textbf{52.88}    \\ 
        
\bottomrule[1.5pt]

\end{tabular}
}
\end{table*}

\subsection{HQ-SAM for Spatial Refinement}
Since SAM has shown its great ability in segmenting objects, it could serve as a spatial refiner for better results. Specifically, in this report, we adopt HQ-SAM \cite{ke2024segment} with ViT-L as our mask refiner. Given the predicted result from MUTR of each clip, we first determine the coordinates of the bounding box by selecting the maximum and minimum horizontal and vertical coordinates of the points along the boundary of the mask. Next, we uniformly sample 10 coordinates within the predicted mask as positive points and 5 coordinates out of the mask but within the bounding box as negative points. The sampled points are then fed into the mask decoder of HQ-SAM as prompts to generate the refined masks.

\subsection{Instance Retrieval Model}
Due to the limited number of input frames for training a RVOS model, the existing pipelines face difficulty generating referred masks under long sequences. However, current video instance segmentation (VIS) methods have shown great potential handling long sequences which could help dealing the MeViS task.
Thus, to fully utilize the capabilities of VIS methods for long sequence understanding, we employ a classification model which predict the valid masks sequence under the language expression from candidates generated by VIS model. Specifically, we choose DVIS \cite{zhang2023dvis} to generate the candidate masks with long frame length. The classification follows a simple architecture with Swin-Large and RoBERTa serving as vision and language backbone, respectively. The corresponding vision features are fed into a standard cross-attention module as query with language features as key and value. The obtained features are consequently averaging pooled at the candidate mask level, following a one-hot classifier to obtain the valid mask sequence result under present language expression. 

\subsection{Fusion Strategy}
We find that predicted results of instance retrieval model achieved better temporal consistency. Therefore, we design a fusion strategy to fuse predicted results from two models both frame-level and instance-level. First, we filter results of MUTR-based model with noise and retrieve instance from 
results of instance retrieval model utilizing IOU in a frame-independent manner. Then, we utilize the frame-level fusion results to retrieve the instance from the whole video utilizing IOU. 

\begin{table}[t]
\centering
\caption{\textbf{Ablation Experiment Results on MeViS Validation Set about Fusion Strategies.} We ablate fusion strategies about frame-level fusion and instance-level fusion.}
\label{tab:fusion_ablation}
\vspace{5pt}
\resizebox{1.0\linewidth}{!}{
\begin{tabular}{c|c|ccc}
\toprule[1.5pt]
Index &Method     & $\mathcal{J}\&\mathcal{F}$   & $\mathcal{J}$   & $\mathcal{F}$      \\ 
\midrule 
 1 &Instance retrieval  &49.23   &45.34  &53.12 \\
 2 &MUTR-based &50.27   &47.67  &52.88 \\
 3 & (1 \& 2) + frame-level  &52.42   & 49.10  &55.75 \\
 4 & (1 \& 3) + instance-level  &\cellcolor{gray!25}\textbf{52.67}   &\cellcolor{gray!25}\textbf{49.08}  &\cellcolor{gray!25}\textbf{56.26} \\
        
\bottomrule[1.5pt]

\end{tabular}
}
\end{table}

\section{Experiments}
\subsection{Dataset and Metrics}
\noindent{\textbf{Datasets.}} We train and evaluate our solution on MeViS, a large-scale dataset for referring video segmentation. It contains 2,006 videos with 28,570 language expressions in total. These videos are divided into 1,662 videos for training, 50 videos for offline evaluation, 140 videos for online evaluation, and other videos for competition.

\noindent{\textbf{Metrics.}} We adopt standard evaluation metrics for MeViS: region similarity ($\mathcal{J}$), contour accuracy ($\mathcal{F}$), and their average value ($\mathcal{J}\&\mathcal{F}$).




\subsection{Implement Details}
In our solution, we employ global sampling method, which divides the entire video into a set of segments and samples one frame randomly in each segment, allowing model to access frames across the whole video. We adopt the pre-trained weights of MUTR as initialization and fine-tune model on MeViS. The number of sampling frames is 5. The batch size is 1 and the accumulation step is 2. 

\subsection{Ablation Experiments}
To validate the effectiveness of our solution, we conduct simple ablation experiments. Experiment results are shown in \cref{tab:ablation} and \cref{tab:fusion_ablation}. It is noted that utilizing HQ-SAM for refinement brings an improvement on $\mathcal{J}$ while a drop about $\mathcal{F}$. However, utilizing HQ-SAM for refinement still brings an improvement on $\mathcal{J}\&\mathcal{F}$. 
When we combine all methods, our solution achieves the best performance 52.67 $\mathcal{J}\&\mathcal{F}$.

\subsection{Competition Results}
Finally, we submit our best solution and achieve 60.36 $\mathcal{J}\&\mathcal{F}$ ( 56.88 $\mathcal{J}$ and 63.85 $\mathcal{F}$ ) on test phase, which secures the final ranking of the 3rd place in the 6-th LSVOS Challenge RVOS Track.

{\small
\bibliographystyle{ieee_fullname}
\bibliography{egbib}
}

\end{document}